\documentclass[conference]{IEEEtran}
\IEEEoverridecommandlockouts
\usepackage{cite}
\usepackage{amsmath,amssymb,amsfonts}
\usepackage{graphicx}
\usepackage{textcomp}
\usepackage{xcolor}
\usepackage{bm}
\usepackage{makecell}
\usepackage{tabu, booktabs}
\usepackage{soul,color}

\usepackage[utf8]{inputenc}
\usepackage{bm}
\usepackage{color}

\usepackage{bbding}
\usepackage{wasysym}
\usepackage{algorithm}
\usepackage[noend]{algpseudocode}
\usepackage{multirow}
\usepackage{mathrsfs}
 
\usepackage{makecell}
\usepackage{tabu, booktabs}
\usepackage{soul,color}
\usepackage{caption}
\usepackage{subcaption}

\usepackage{lipsum} 
\usepackage{enumitem}
\newlist{researchquestions}{enumerate}{1}
\setlist[researchquestions]{label*=\textbf{RQ\arabic*}}

\usepackage{framed}

\usepackage{algorithm}
\usepackage[noend]{algpseudocode}
\def\BibTeX{{\rm B\kern-.05em{\sc i\kern-.025em b}\kern-.08em
    T\kern-.1667em\lower.7ex\hbox{E}\kern-.125emX}}
\begin{document}

\title{Adapting Membership Inference Attacks to GNN for Graph Classification: Approaches and Implications}


\author{\IEEEauthorblockN{Bang Wu}
\IEEEauthorblockA{\textit{Monash University} \\
Melbourne, Australia \\
bang.wu@monash.edu}
\and
\IEEEauthorblockN{Xiangwen Yang}
\IEEEauthorblockA{\textit{Monash University} \\
Melbourne, Australia \\
wayne.yang@monash.edu}
\and
\IEEEauthorblockN{Shirui Pan}
\IEEEauthorblockA{\textit{Monash University} \\
Melbourne, Australia \\
shirui.pan@monash.edu}
\and
\IEEEauthorblockN{Xingliang Yuan}
\IEEEauthorblockA{\textit{Monash University} \\
Melbourne, Australia \\
xingliang.yuan@monash.edu}
}

\maketitle

\footnotetext[1]{The short version of this paper has been published in the IEEE International Conference on Data Mining (ICDM) 2021.}

\begin{abstract}
Graph Neural Networks (GNNs) are widely adopted to analyse non-Euclidean data, such as chemical networks, brain networks, and social networks, modelling complex relationships and interdependency between objects. 
Recently, Membership Inference Attack (MIA) against GNNs raises severe privacy concerns, where training data can be leaked from trained GNN models. 
However, prior studies focus on inferring the membership of only the components in a graph, e.g., an individual node or edge. How to infer the membership of an entire graph record is yet to be explored. 

In this paper, we take the first step in MIA against GNNs for graph-level classification. 
Our objective is to infer whether a graph sample has been used for training a GNN model. We present and implement two types of attacks, i.e., training-based attacks and threshold-based attacks from different adversarial capabilities. 
We perform comprehensive experiments to evaluate our attacks in seven real-world datasets using five representative GNN models. %
Both our attacks are shown effective and can achieve high performance, i.e., reaching over $0.7$ attack F1 scores in most cases\footnote[2]{The code and data used in the paper are released at \\
https://github.com/TrustworthyGNN/MIA-GNN/}. 
Furthermore, we analyse the implications behind the MIA against GNNs. 
%
Our findings confirm that GNNs can be even more vulnerable to MIA than the models with non-graph structures. %
And unlike the node-level classifier, MIAs on graph-level classification tasks are more co-related with the overfitting level of GNNs rather than the statistic property of their training graphs. 
\end{abstract}



\section{Introduction}

Graph-structured data, with powerful capability presenting wealthy information of real-world data, has been broadly used in various applications such as social networks, medical screening, and chemical property prediction~\cite{XuHXXXY20,WangYDHLN19}. 
To analyse rich graph data, deep learning methods have drawn much attention recently. 
Among them, a family of algorithms, called Graph Neural Networks (GNNs), have achieved state-of-the-art performance by generalising neural networks for graphs~\cite{wu2020comprehensive,liu2021graph}. 

Despite the great power of GNNs, privacy concerns about information exposure in GNNs have been raised in sensitive applications~\cite{ChenYZF20,ZhuLLBZZL20}. 
The graph data for model training is commonly considered as a private property of data owners. 
For example, chemical and biomedical networks carefully collected from highly consuming experiments are deemed as proprietary assets of companies. 
%
In addition, some graph data may contain sensitive attributes, e.g., diagnosis of customers' behaviours. 
The leakage and abuse of such data may result in serious issues. 

Prior studies show that many deep learning methods are vulnerable to a severe privacy attack named Membership Inference Attack (MIA)~\cite{ChenYZF20,PangZJLW20,263820,GongL18,abs-2007-14321,demystifying19,Salem0HBF019,ShokriSSS17}.
Given access to a model, the attacker can infer whether an arbitrary data record has been used during the training period of this model. 
As an early study, Shokri \textit{et al.}~\cite{ShokriSSS17} have demonstrated that the attacker can propose a binary classifier identifying the membership of a record based on its posterior, i.e., the confidence value predicted by the target model. 
Since the ordinary neural networks often generate different posteriors for the member or non-member inputs, such an attack achieves high success rates and poses significant threats to the privacy of the sensitive training data.

In the domain of GNNs, recent studies~\cite{263820,abs-2101-06570,GongL18} start to explore the feasibility of MIA over node-level GNN models.
Specifically, node membership inference attacks~\cite{he2021node,abs-2101-06570} can infer whether a given node has been used during the training of a target GNN model. Some other inference attacks ~\cite{263820} target at connectives and predict whether a specific pair of nodes are connected in the training graph. 
%
Note that those works only infer the membership of a component in the graph.
Therefore, how to realise GNN MIAs in graph classification and how an entire graph record can be leaked by the GNNs are yet to be explored. 

In this paper, we aim to thoroughly investigate the GNN MIAs in graph-level classification attacks. 
As aforementioned, the first question is how to realise such attacks in graph classification. 
%
In prior inference attacks against node-level GNNs, attackers infer the membership for only a node's attributes or connectivity between two nodes based on the inherent property of graph data. 
%
For example, the membership inference of the connectivity between two nodes is based on the similarity of their attributes or predictions from the target model~\cite{263820}. 
Therefore, such inference is implemented by utilising the strong correlation between connected nodes, which is common in graph-structure data.
We observe that the above intuition does not appear to be extended to graph-level GNN MIAs for inferring the membership of individual graph records.  

The second question we aim to tackle is how vulnerable GNN models are to the MIAs. 
Specifically, \textit{1) What factors and how do they impact the performance of membership inference attacks? }
According to previous studies~\cite{ShokriSSS17,YeomGFJ18} in DNN models, overfitting is the most significant cause for the MIAs, while studies~\cite{he2021node,abs-2101-06570} in node-level GNNs show that the graph property is also a significant contributor.
In this paper, we will explore how GNN models memorise the training records and perform differently under divergent overfitting levels and classification tasks of various graph data. 
\textit{2) How is the transferability of the MIAs on GNNs? }
Most of the MIAs on DNN are shown to have strong transferability, i.e., the classifier identifying the membership of models can also infer the membership for the model trained with different domain data~\cite{ShokriSSS17}. 
The transferring attacks with enhanced generalisability pose larger privacy threats. 
Therefore, the transferability of MIA on GNNs needs to be investigated.

\noindent \textbf{Contributions.} We present the first MIA attacks on graph-level GNN tasks in this work.
%
To address the first question, we propose two types of attacks, i.e., \textit{training-based attacks} and \textit{threshold-based attacks} based on the different capabilities of the attackers. 
Both attacks can effectively identify the membership of a graph record regarding a target GNN model. 
Specifically, the attacker issues a query to the target model and receives confidence scores as a response. 
After that, for the training-based attack, a binary classifier is constructed to infer the membership based on the confidence scores. 
For the threshold-based attack, the attacker first extracts a confidence value, e.g., the highest confidence score, and then infers the membership by comparing this value to a threshold. 
Intuitively, since the confidence scores are different for member/non-member inputs, our attacks can identify their memberships in high confidence. 
We perform a set of experiments to assess our MIAs among different aggregation methods and training data for both our proposed training-based and threshold-based attacks. 

To further investigate how vulnerable the GNN models are to the MIAs, we comprehensively measure the effectiveness of our attacks under various experimental settings from two perspectives: GNN methods and training datasets. 
In particular, we evaluate our attacks against GNN models with different overfitting levels by adjusting the hyperparameters during the training, such as the training epoch and model architecture. 
Meanwhile, we perform another set of experiments to assess the co-relations between the statistic property of the training graph and the performance of our MIAs. 
We comprehensively compare our graph-level MIAs to other node-level attacks and analyse the difference and implications behind them. 

The contributions of our work are summarised as follows:

\begin{itemize}
    \item We propose \textit{the first} GNN membership inference attacks for the graph-level classification tasks with black-box access to target models.

    \item We propose two types of attacks to infer the membership of an arbitrary graph record from learning-based and threshold-based approaches, respectively. We present how to build the training dataset for learning-based attacks and how to choose the confidence value metric for the threshold-based attacks. 

    \item We evaluate our attacks in seven real-world datasets from different domains using five representative GNN methods and conduct extensive evaluations for the transferability of our attacks. We thoroughly analyse the factors which impact the attack performance and reveal the implications of MIAs against GNNs on both node-level and graph-level classification tasks. 
\end{itemize}

\noindent \textbf{Highlights of our key findings.}
Particularly, we highlight the significant findings corresponding to the aforementioned research questions as follows. 
\begin{itemize}
    \item GNNs are vulnerable to the membership inference attacks and can be even more vulnerable than the ML models with non-graph structures in certain applications. 
    \item Overfitting is the most significant factor for both training-based or threshold-based attacks, which is consistent with the observations of prior MIAs in DNN but different from prior MIAs in node-level GNN models. 
    \item The training-based attacks have strong transferability among multiple GNN types and shadow datasets, while the threshold-based attacks achieve higher attack performance but poorer transferability. 
\end{itemize}

\section{Related Work}


Recent studies~\cite{ShokriSSS17,YeomGFJ18,Salem0HBF019,demystifying19,LeinoF20} have shown that attackers can infer the training records of various machine learning models via MIAs, and achieve outperformed attack success rate and precision. 
Shokri \textit{et al.}~\cite{ShokriSSS17} first present the MIA against ML models by constructing a binary classifier. 
The classifier can infer the membership based on the posterior of a record. 
Leino \textit{et al.}~\cite{LeinoF20} present an effective MIA with high precision under white-box settings. 
Chen \textit{et al.}~\cite{ChenYZF20} extend the MIA to generative models and develop a comprehensive framework for their attacks. 
Salem \textit{et al.}~\cite{Salem0HBF019} propose three MIAs for the attacker with different capabilities, and show that their attacks can be broadly applicable at low cost. 
Christopher \textit{et al.}~\cite{abs-2007-14321} investigate label-only attacks, which can infer the membership based on only the prediction label of the records. 
However, all the above-mentioned works show the successful MIAs on the machine learning model trained on Euclidean space. 
In this paper, we study MIA against graph-level classifiers via graph neural networks (GNNs), which is a family of ML models trained on graph structure data. 

There are also several preliminary researches~\cite{he2021node,abs-2101-06570,263820} on the MIAs against node-level classifiers. 
However, these node-level MIAs only utilise the information of the sub-graph around the target node. 
For example, He \textit{et al.}~\cite{263820} derive the membership of a link by the variant distance between the posteriors of only the corresponding two nodes.  
Their other work~\cite{he2021node} and Olatunji \textit{et al.}~\cite{abs-2101-06570} which identify the membership of a specific node, infer the final results based on the combination of the posterior of the target nodes and its 2-hop neighbour. 
All of them do not consider the embedding of the entire graph which falls short of showing the vulnerability of GNNs on graph-level. 
Recently, Zhang \textit{et al.}~\cite{InferAttack22} infer the basic graph properties and the membership of a sub-graph based on graph embedding, but not the membership of the entire graph.
Therefore, how graph-level GNN classifiers are vulnerable to MIAs and what is the insights behind them are still yet to be explored.
%

\section{Problem Formalisation}

\label{sec:problem}
This section presents the formal definition of the proposed MIAs, and introduces the attack assumptions. 

\subsection{Problem Definition}

\noindent \textit{Definition 2.1.} (Graph Classification Model:) Graph classification aims to predict a categorical class for a given graph. 
Specifically, with a set of graphs $G=\{g_1, g_2, \cdots, g_n\}$ where each individual $g_i$ has a class $y_i$ in a set of label $Y$, a graph classification model is a mapping $f_{\theta}(.) : g_i \rightarrow y_i$ which infers the label of the graph. 

GNN based graph classification is a fundamental analytics task and serves many applications such as fraud detection \cite{HuangXYKAGM13,GongFM14,abs-1908-02591,WangJG19},
malware detection \cite{KongY13,NikolopoulosP17,HassenC17,YanYJ19}, and healthcare \cite{abs-1709-03741}. It has attracted increasing attention due to its high prediction accuracy, and has become the standard toolkit for analysing and learning from data on graphs. 
Here, we define the graph classification model as the victim model and target its training data membership's confidentiality.
The training dataset $G=\{(g_1,y_1),(g_2,y_2),...,(g_n,y_n)\}$ is a set of labeled graphs with ground truth labels. 
The objective function of the graph classifier can be formulated as:
\begin{equation}
    \min_{\theta} \sum_{g_i \in G, y_i \in Y} L(f_{\theta}(g_i),y_i),
\end{equation}
\noindent where $L(.)$ is the loss function of the graph prediction. 


\noindent \textit{Definition 2.2.} (Membership Inference Attacks in GNNs:) Given a GNN model $f_{\theta}(.)$, it is trained on a labeled graph set $G_{Train}=(G,Y)$ for a graph-level classification task.
The membership inference attack attempts to infer whether a specific graph $g_i$ is in the target training dataset $G$.
Formally, 
\begin{equation}
    \mathcal{A}: (g_i,f_{\theta}(.)) \rightarrow \{0,1\},
\end{equation}
\noindent where $\mathcal{A}$ represents the attack model which outputs 1 if the record $g_i$ is in the training set of $f_{\theta}(.)$, and outputs 0 otherwise.

\subsection{Attack Assumptions}\label{AA}
This section introduces the detailed attack scenarios by explaining the attacker's capability and background knowledge.

\noindent \textbf{Black-box settings.} We assume that the attacker can only obtain black-box access to the target models, which is the most common and realistic setting for the adversarial knowledge~\cite{PangZJLW20,JiaSBZG19}.
In the black-box setting, the attacker has access to neither the parameters of the target models nor the internal representations during the inference. 
Only the model queries (model outputs of a chosen input) are accessible. Specifically, the well-trained models are normally deployed in the MLaaS, which only provides model queries to other users through an API. 
The black-box attacker attempts to exploit the difference between the prediction of their chosen queries and infer the membership of the records. 

\noindent \textbf{Attacker's background knowledge. }
Even though the attacker has only black-box access to the model parameters and internal representations, he can manage to obtain some public information. 
\begin{itemize}
    \item \textbf{Shadow dataset. }
    A Shadow dataset can be a dataset with the same domain as the target model training dataset. 
    Since the target model's task is commonly public knowledge, the attacker can gather data and build the shadow dataset. 
    In this paper, we also relax this assumption in the case that the shadow dataset comes from a different domain. 
    \item \textbf{Training knowledge. }
    We assume the attacker knows the type of GNNs and how a GNN was trained (i.e., the training algorithm). This assumption is reasonable since many packages such as Deep Graph Library (DGL) \cite{wang2019dgl} provides implementation of GNN models and the assumption is consistent with prior MIAs~\cite{ShokriSSS17,Salem0HBF019}. 
    We will further relax this assumption and discuss our attacks when the attacker does not know the type of GNNs.
\end{itemize}

\section{The Proposed Attacks}
\label{sec:method}
Fundamentally, MIAs exploit different responses between the member and non-member inputs.
As discussed in Section~\ref{sec:problem}, in the black-box settings, the attacker can only obtain the responses, and the intuition of applying MIAs is to analyse the responses from the target model and identify the membership. There are two main methods of MIAs: \textit{training-based} and \textit{non-training-based} approaches. Here, we design our attacks from these two approaches, respectively.

\subsection{Training-based Attacks}
\noindent \textbf{Overviews. }
The idea of training-based attacks is to train an attack model to classify the membership of the input.
This attack model will be a binary classifier whose output is ``in" or ``out" corresponding to whether one record has been or not been used during the target model training.
Therefore, we first propose to construct a training dataset for the attack model.
Meanwhile, supervised training is adopted to obtain a more accurate binary classifier, as it is better to build a training dataset whose records have been labelled as ``in" or ``out". 
With only black-box access to the target model, the attacker has no idea about the target model's training dataset, as he cannot get the membership label of a record. 
To construct the attack model, the attacker needs to build a surrogate model based on the background knowledge. 

\noindent \textbf{Construction. }
Algorithm~\ref{alg:attack_model} shows how to construct the attack model and the detailed steps are listed as followed:

\begin{algorithm}[!t]
\caption{Algorithm for Attack Model Training}\label{alg:attack_model}
\begin{flushleft}
\hspace*{\algorithmicindent}\textbf{Input:} \\
\hspace*{\algorithmicindent} Shadow Dataset $G_s$ \\
\hspace*{\algorithmicindent}\textbf{Output:} \\
\hspace*{\algorithmicindent} Attack Model $f_{attack}(.)$. \\
\end{flushleft}
\begin{algorithmic}[1]
\State Split $G_s$ to $G_{member}$ and $G_{non\_member}$
\State Train Shadow Model $f_{shadow}(.)$ on $G_{member}$
\State $V_{attack} = \emptyset$
\For {$g_i$ in $G_s$}
    \If{$g_i$ in $G_{member}$}
        \State $V_{attack} = V_{attack} \cup \{(f_{shadow}(g_i),``in")\}$
    \EndIf
    \If{$g_i$ in $G_{non\_member}$}
        \State $V_{attack} = V_{attack} \cup \{(f_{shadow}(g_i),``out")\}$
    \EndIf
\EndFor
\State Train Attack Model $f_{attack}(.)$ on $V_{attack}$
\end{algorithmic}
\end{algorithm}

\begin{itemize}
    \item \textit{Step 1. Shadow Datasets Processing. }
    The attacker gathers the dataset and splits it into two parts. 
    One for shadow model training is the shadow model's membership set, and the other one is the non-member set. 
    
    \item \textit{Step 2. Shadow Model Training. }
    The attacker generates a surrogate model to mimic the prediction behaviour of the target model. 
    Only the membership set constructed in the first step is used to train the model. 
    Since the shadow dataset is of the same or similar domain as the target model, it is expected to behave similarly to the target model.

    \item \textit{Step 3. Confidence Scores Gathering. }
     The attacker then feeds both membership and non-membership datasets into the shadow model and obtains their confidence scores. 
    These confidence scores can be labelled as ``in" or ``out" corresponding to the dataset they are from. 
    
    \item \textit{Step 4. Attack Model Training. }
    The attacker finally trains the attack model to identify the membership of a record based on its confidence score. 
    The training data for the attack model is the confidence score set generated in step 3. 
  
\end{itemize}

To apply our attack on an arbitrary record, the attacker first issues a query to the target model and obtains the confidence score of the record. 
Then the attack model can infer the membership of the record based on its confidence score. 
In general, if the shadow model performs similarly to the target, the attack model can correctly identify the record's membership. 

\noindent  \textbf{Remarks.} If the shadow and target datasets are in the same domain, the confidence scores for them will have equal dimensions. However, in reality, the attacker may not have knowledge of the data domain. Thus, the obtained datasets may come from other domains which result in different class numbers, i.e., the dimensions of confidence scores are different to the target one. 
To address the issue, we propose to reduce the confidence scores with higher dimensions to be consistent with the others. Only the most significant values are kept during the reduction. That is, we select the top $k$ values in terms of the confidence scores and construct the new confidence score with $k$ dimensions. 

\subsection{Threshold-based Attacks}
\label{subsec:dimension}
We then present how to apply the attacks without training the attack models. 
In some applications, the attacker may not have appropriate resources for expensive model training during membership inference. Instead, a non-training based approach with a much lower cost can be applied.

To identify the membership of a record only based on the output posteriors, the idea is that the model is more confident with the inputs it memorises. 
Thus, the input with comparably higher confidence can be considered as membership. 
To evaluate how the model is confident with the input, we propose to employ three different evaluation metrics. 
Detailed information is shown in Table~\ref{tab:threshold_metrics}. 

\begin{table}[t]
\normalsize
    \centering
    \begin{tabular}{c|c|c}
    \hline
        \multicolumn{2}{c|}{Metrics} & Definition \\
    \hline
        \multicolumn{2}{c|}{Highest confidence score} & $max({x_1,x_2,...,x_n})$ \\
    \hline
        \multirow{2}{*}{Loss} & MSE & $\sqrt{\sum_i (x_i-y_i)}$ \\
         & Cross-Entropy & $-log(\frac{exp(max(x_i))}{\sum_j exp(x_j)})$ \\
    \hline
        \multirow{2}{*}{Distance} & Cityblock & $\sum_i |x_i-y_i|$\\
         & Canberra & $\sum_i \frac{|x_i-y_i|}{|x_i+y_i|}$ \\
    \hline
    \end{tabular}
    \caption{Confidence metrics. $x_i$ represents the i-th component of confidence vector $x$. $y_i$ represents the i-th component of a one-hot vector $y$ corresponding to the predicted label. }
    \label{tab:threshold_metrics}
\end{table}

\noindent \textbf{Confidence via confidence score. }
The label with the highest confidence score is considered the final prediction label. 
A higher score represents higher confidence in the model regarding the inference. 
Therefore, it is common to use the highest confidence score to represent how the model is confident with the input.

\noindent \textbf{Confidence via loss.}
It is also reasonable to use the loss function as a metric for the confidence of the records. 
In particular, since the model is trained to reduce the loss of the training data, the loss of the memberships is expected to be lower than the non-memberships. 
Therefore, we calculate the loss of a record, and a smaller loss value means higher confidence from the target models. 

In the proposed attacks, two popular loss functions, Mean Squared Error (MSE) and Cross-entropy are used as the confidence metrics. 
Note that, the calculation of the loss during the training requires the ground truth labels of the data. %
In our attack settings, the attacker cannot obtain these ground truth labels. Nevertheless, since the target models are often well-trained, we can use the predictions as their labels. 
So, we can still calculate the loss values and apply the threshold-based attacks.

\noindent \textbf{Confidence via distance. }
Besides using some popular loss functions, we propose to measure the confidence via distance. 
Similar to the calculation of loss, we evaluate the distance between the confidence vectors and the one-hot output vectors corresponding to the ground truth labels. 
Since some distances are positively related to the previous metrics, e.g., MSE is the same as the Euclidean distance, we select Cityblock distance (also called Manhattan distance) and Canberra distance as our confidence metrics.

\noindent \textbf{Threshold selection. }
 Given the confidence values, the attacker can infer the membership via a threshold value. 
The threshold value can be chosen by considering the requirements of the attacks as prior work~\cite{Salem0HBF019}. 
Specifically, the higher threshold can be selected when the attacker focuses more on precision. 
In contrast, he can choose a lower threshold if he concentrates more on recall. 

\section{Evaluation and Analysis}
\label{sec:exp}

Our evaluation aims at answering the following research questions:
\begin{researchquestions}

\item \textit{Whether and how different types of GNNs can be vulnerable to MIAs? }

\item \textit{How is the transferability of MIAs? 
how is the attack performance affected if the attacker obtains less knowledge of the target model? }



\item \textit{What are the factors that impact the performance of MIAs? 
Is the overfitting only factor that affects the performance of attacks in GNNs?  }

\item \textit{Does the target model performance (e.g. overfitting level) or the target dataset property (e.g. statistics of the graph) affect the attack performance more? What is the difference between the MIAs on graph-level classification tasks and other graph-based MIAs? }




\end{researchquestions}

To answer RQ1, we apply our attacks on various types of GNN models trained by different graph data, and then compare them with the baseline attacks on MLP. 
To answer RQ2, we apply the attacks with different adversarial background knowledge and evaluate their effects.
To answer RQ3, we adjust several parameters to analyse their impacts on attack performance. 
To answer RQ4, we discuss the empirical results under different settings and compare our observations to prior studies. 

In this section, we first introduce the experimental settings for our attacks. 
Then, we present the results for both learning-based and threshold-based attacks regarding the above research questions and further discuss and compare our observations with others.  


\subsection{Experimental Setup}

\begin{table}[t]
    \small
    \centering
    \begin{tabular}{c|cccc}
    \hline
        Dataset & \#Graphs & \#Classes & Avg. Nodes & Avg. Edges \\
        \hline
         PROTEIN\_full & 1113 & 2 & 39.06 & 62.14 \\
         DD & 1178 & 2 & 284.32 & 715.66 \\
         ENZYMES & 600 & 6 & 32.63 & 62.14 \\
         CIFAR10 & 60K & 10 & 117.63 & 941.07 \\
         MNIST & 70K & 10 & 70.57 & 564.53 \\ 
         OGBG-PPA & 158K & 37 & 243.4 & 2266.4 \\ %
         NCI & 4110 & 2 & 29.87 & 32.30 \\
         \hline
    \end{tabular}
    \caption{Summary statistics of seven datasets. }
    \label{tab:dataset}
\end{table}

\begin{table*}[t]
\normalsize
    \centering
    \begin{tabular}{c|c|ccc|ccc}
    \hline
        Dataset & Model & \makecell[c]{Training \\ Accuracy} & \makecell[c]{Testing \\ Accuracy} & \makecell[c]{Train-test \\ Gap} & \makecell[c]{Attack \\ Precision} & \makecell[c]{Attack \\ Recall} & \makecell[c]{Attack \\ F1 Score} \\
    \hline
        \multirow{5}{*}{PROTEIN\_ful} & GateGCN & 0.990 & 0.710 & 0.280 & 0.602(0.070) & 0.588(0.058) & 0.595(0.063) \\
                      & GCN & 1.000 & 0.688 & 0.313 & \bm{$0.745(0.055)$} & \bm{$0.607(0.046)$} & \bm{$0.668(0.043)$} \\
                      & GIN & 0.730 & 0.690 & 0.040 & 0.563(0.078) & 0.559(0.074) & 0.561(0.075) \\
                      & GAT & 1.000 & 0.660 & 0.340 & 0.678(0.094) & 0.590(0.047) & 0.630(0.066) \\
                      & MLP(baseline) & 0.990 & 0.660 & 0.340 & 0.691(0.086) & 0.602(0.057) & 0.642(0.065) \\
    \hline
                   \multirow{5}{*}{DD} & GateGCN & 1.000 & 0.630 & 0.370 & \bm{$0.890(0.029)$} & \bm{$0.881(0.035)$} & \bm{$0.885(0.030)$} \\
                      & GCN & 1.000 & 0.630 & 0.370 & 0.793(0.065) & 0.682(0.059) & 0.733(0.057) \\
                      & GIN & 1.000 & 0.610 & 0.390 & 0.600(0.050) & 0.594(0.048) & 0.597(0.051) \\
                      & GAT & 1.000 & 0.670 & 0.330 & 0.831(0.035) & 0.758(0.043) & 0.792(0.036) \\
                      & MLP(baseline) & 0.780 & 0.650 & 0.130 & 0.553(0.054) & 0.551(0.054) & 0.552(0.054) \\
    \hline
              \multirow{5}{*}{ENZYMES} & GateGCN & 1.000 & 0.550 & 0.450 & \bm{$0.865(0.083)$} & 0.715(0.072) & \bm{$0.782(0.072)$} \\
                      & GCN & 1.000 & 0.520 & 0.480 & 0.843(0.101) & 0.725(0.088) & 0.778(0.088) \\
                      & GIN & 0.940 & 0.480 & 0.460 & 0.593(0.071) & 0.592(0.063) & 0.592(0.067) \\
                      & GAT & 1.000 & 0.530 & 0.470 & 0.817(0.044) & \bm{$0.736(0.051)$} & 0.774(0.043) \\
                      & MLP(baseline) & 1.000 & 0.380 & 0.620 & 0.821(0.072) & 0.625(0.063) & 0.707(0.058) \\
    \hline
              \multirow{6}{*}{CIFAR10} & GateGCN & 1.000 & 0.476 & 0.524 & 0.877(0.011) & \bm{$0.843(0.009)$} & 0.859(0.010) \\
                      & GCN & 0.995 & 0.363 & 0.632 & 0.760(0.014) & 0.748(0.012) & 0.754(0.013)\\
                      & GIN & 0.984 & 0.319 & 0.666 & 0.601(0.013) & 0.599(0.013) & 0.600(0.013) \\
                      & GAT & 0.989 & 0.416 & 0.573 & 0.693(0.013) & 0.672(0.012) & 0.682(0.012) \\
                      & GraphSAGE & 1.000 & 0.494 & 0.506 & \bm{$0.881(0.009)$} & 0.841(0.007) & \bm{$0.860(0.008)$} \\
                      & MLP(baseline) & 1.000 & 0.354 & 0.646 & 0.768(0.013) & 0.680(0.009) & 0.721(0.010) \\
    \hline
                \multirow{6}{*}{MNIST} & GateGCN & 1.000 & 0.918 & 0.082 & 0.673(0.068) & 0.585(0.009) & 0.625(0.031) \\
                      & GCN & 1.000 & 0.805 & 0.195 & \bm{$0.803(0.010)$} & \bm{$0.720(0.007)$} & \bm{$0.759(0.008)$} \\
                      & GIN & 0.984 & 0.755 & 0.228 & 0.547(0.010) & 0.546(0.010) & 0.547(0.010) \\
                      & GAT & 1.000 & 0.888 & 0.112 & 0.709(0.071) & 0.596(0.009) & 0.646(0.033) \\
                      & GraphSAGE & 1.000 & 0.910 & 0.090 & 0.762(0.027) & 0.636(0.005) & 0.693(0.013) \\
                      & MLP(baseline) & 1.000 & 0.856 & 0.144 & 0.666(0.058) & 0.567(0.009) & 0.612(0.028) \\
    \hline
            OGBG\_PPA & DeepGCN & 1.000 & 0.588 & 0.412 & 0.843(0.014) & 0.810(0.010) & 0.826(0.011) \\
    \hline
                  NCI & GCN & 1.000 & 0.622 & 0.378 & 0.689(0.038) & 0.613(0.024) & 0.649(0.029) \\

    \hline
    \end{tabular}
    \caption{Accuracy of the target models for different datasets and the corresponding performance on training-based attacks.}
    \label{tab:overall}
    \vspace{-5pt}
\end{table*}

\begin{table*}[t]
\normalsize
    \centering
    \begin{tabular}{c|c|c|cc|cc}
    \hline
        Dataset & Model & \makecell[c]{Highest \\ Confidence Score} & \makecell[c]{Cross-Entropy} & \makecell[c]{MSE} & \makecell[c]{ Cityblock } & \makecell[c]{ Canberra }\\
    \hline
        \multirow{5}{*}{PROTEIN\_ful} & GateGCN & 0.501(0.070) & \underline{\bm{$0.678(0.021)$}} & 0.326(0.029) & 0.400(0.061) & 0.399(0.061) \\
                      & GCN & 0.533(0.148) & \underline{\bm{$0.690(0.028)$}} & 0.232(0.019) & 0.263(0.060) & 0.263(0.049) \\
                      & GIN & 0.297(0.010) & \underline{\bm{$0.665(0.005)$}} & 0.107(0.001) & 0.144(0.009) & 0.144(0.008) \\
                      & GAT & 0.544(0.026) & \underline{\bm{$0.677(0.028)$}} & 0.284(0.065) & 0.241(0.055) & 0.284(0.065) \\
                      & MLP & 0.544(0.012) & \underline{\bm{$0.701(0.027)$}} & 0.292(0.020) & 0.285(0.098) & 0.285(0.084) \\
    \hline
                   \multirow{5}{*}{DD} & GateGCN & 0.406(0.340) & \bm{$0.845(0.049)$} & 0.366(0.017) & 0.426(0.030) & 0.426(0.079) \\
                      & GCN & 0.666(0.006) & \underline{\bm{$0.771(0.001)$}} & 0.209(0.027) & 0.191(0.063) & 0.191(0.045) \\
                      & GIN & 0.044(0.009) & \underline{\bm{$0.673(0.002)$}} & 0.000(0.000) & 0.015(0.006) & 0.015(0.005) \\
                      & GAT & 0.724(0.007) & \underline{\bm{$0.793(0.007)$}} & 0.584(0.002) & 0.194(0.004) & 0.194(0.006) \\
                      & MLP & 0.046(0.029) & $0.533(0.014)$ & 0.507(0.011) & \underline{\bm{$0.598(0.009)$}} & \underline{\bm{$0.598(0.008)$}} \\
    \hline
              \multirow{5}{*}{ENZYMES} & GateGCN & \underline{$0.844(0.052)$} & \underline{\bm{$0.848(0.038)$}} & 0.581(0.034) & 0.598(0.075) & 0.598(0.070) \\
                      & GCN & \underline{$0.812(0.030)$} & \underline{\bm{$0.854(0.019)$}} & 0.578(0.009) & 0.684(0.000) & 0.684(0.000) \\
                      & GIN & 0.011(0.015) & \underline{\bm{$0.668(0.004)$}} & 0.590(0.024) & 0.579(0.009) & 0.579(0.036) \\
                      & GAT & \underline{$0.841(0.007)$} & \underline{\bm{$0.854(0.009)$}} & 0.584(0.002) & 0.682(0.004) & 0.682(0.003) \\
                      & MLP & \underline{$0.715(0.034)$} & \underline{\bm{$0.807(0.011)$}} & 0.576(0.027) & 0.581(0.008) & 0.581(0.006) \\
    \hline
              \multirow{6}{*}{CIFAR10} & GateGCN & 0.844(0.011) & \underline{\bm{$0.860(0.007)$}} & 0.542(0.0003) & 0.561(0.013) & 0.561(0.011) \\
                      & GCN & 0.688(0.035) & \underline{\bm{$0.759(0.012)$}} & 0.547(0.005) & $0.554(0.013)$ & 0.554(0.015)\\
                      & GIN & 0.421(0.221) & \underline{\bm{$0.757(0.176)$}} & 0.544(0.016) & 0.567(0.027) & 0.567(0.023) \\
                      & GAT & 0.314(0.161) & \bm{$0.581(0.050)$} & 0.550(0.007) & 0.573(0.020) & 0.573(0.026) \\
                      & GraphSAGE & 0.697(0.154) & \bm{$0.755(0.107)$} & 0.552(0.005) & 0.556(0.001) & 0.556(0.001) \\
                      & MLP & 0.482(0.125) & \bm{$0.661(0.064)$} & 0.542(0.002) & 0.544(0.008) & 0.544(0.011) \\
    \hline
                \multirow{6}{*}{MNIST} & GateGCN & 0.445(0.030) & \underline{\bm{$0.712(0.008)$}} & 0.539(0.005) & 0.575(0.005) & 0.575(0.004) \\
                      & GCN & 0.551(0.077) & \bm{$0.662(0.075)$} & 0.546(0.01) & 0.545(0.010) & 0.545(0.025) \\
                      & GIN & 0.120(0.023) & \underline{\bm{$0.674(0.002)$}} & 0.541(0.001) & \underline{0.587(0.003)} & \underline{0.587(0.009)} \\
                      & GAT & 0.404(0.042) & \underline{\bm{$0.719(0.006)$}} & 0.542(0.002) & 0.577(0.004) & 0.577(0.009) \\
                      & GraphSAGE & 0.517(0.023) & \underline{\bm{$0.737(0.006)$}} & 0.544(0.001) & 0.593(0.001) & 0.583(0.000) \\
                      & MLP & 0.484(0.024) & \underline{\bm{$0.671(0.034)$}} & 0.541(0.001) & 0.580(0.001) & 0.580(0.005) \\
    \hline
            OGBG\_PPA & DeepGCN & 0.717(0.152) & \bm{$0.796(0.088)$} & 0.526(0.005) & 0.578(0.007) & 0.577(0.008) \\
    \hline
    \end{tabular}
    \caption{F1 score of the threshold-based attacks targeting different models for six datasets. The results that are higher than training-based attacks are underlined. The best results are bold. }
    \label{tab:overall_threshold}
\end{table*}

To obtain a comprehensive insight into how GNNs are vulnerable to the MIAs, we evaluate the attacks among various types of GNNs trained by multiple different datasets. 

\noindent \textbf{Datasets.} We use seven real-world datasets from different domains to evaluate our attacks. Specifically, PROTEIN\_full~\cite{abs-2003-00982}, DD~\cite{dobson2003distinguishing}, ENZYMES~\cite{abs-2003-00982} and OGBG-PPA~\cite{SzklarczykGLJWH19} are all protein graph datasets. 
CIFAR10~\cite{tiny_images} and MNIST~\cite{726791} are classical image classification datasets converted into graphs using super-pixels~\cite{AchantaSSLFS12}. 
The graphs in NCI~\cite{abs-2007-08663} represent the atoms and chemical bond connections of the chemical compounds. 
The statistics of the datasets is given in Table~\ref{tab:dataset}. 

Here, we evaluate our attacks by considering the worst-case scenario for the attacker, where no same data in the shadow dataset is compared to the target one. In particular, we equally split them into two equal parts: one is the target models' training dataset, while the other is the shadow dataset obtained by the attackers. 
Note that in practice, it is possible for the attacker to obtain a shadow dataset overlapping with the target dataset. 
In this case, the shadow model will be more similar to the target model and benefit our attacks. 
The purpose of this setting is to show that even in the worst case, our attacks are quite powerful and can still achieve high performance in membership inference.

\noindent \textbf{GNN types.} We evaluate five popular GNNs using these datasets.
For PROTEIN\_full, DD, ENZYMES, we use GCN~\cite{KipfW17}, GateGCN~\cite{abs-1711-07553}, GIN~\cite{XuHLJ19}, GAT~\cite{VelickovicCCRLB18} as the target models. 
And for CIFAR10 and MNIST, we use the above four GNNs as well as GraphSAGE~\cite{HamiltonYL17} for the target models. 
As a common setting in practice, all of these models contain only two layers. 
To investigate the vulnerability of deeper GNNs with more parameters, we further evaluate DeepGCN~\cite{Li0TG19} in OGBG-PPA and NCI. 
We adjust the architectures of these deeper GNNs and demonstrate their impacts. 
We also use MLP as a non-graph-structure baseline model as~\cite{abs-2003-00982} to help evaluate our attacks.
It simply updates the representation for each node independently without considering their neighbours. 
Other parameters, such as the number of the model layers, are set to be the same as GNNs.

\noindent \textbf{Evaluation metrics.} Because membership inference is a binary classification problem, in our experiments, we adopt the attack precision, recall and F1 score as our evaluation metrics to measure the overall performance of our attacks, which are consistent with prior work on MIAs~\cite{ShokriSSS17,YeomGFJ18}. 
We run all our experiments 15 times.

\subsection{Attack Performance}

\subsubsection{Performance Overview}
We first show how different types of GNN models are vulnerable to  both our training-based and threshold-based attacks on different datasets.  

\begin{framed}
    \noindent \textbf{Findings \#1:} 
    \textit{Several popular types of GNN models are all vulnerable to the MIAs on different training data. }
\end{framed}
\noindent \textbf{Training-based attacks.} 
Table~\ref{tab:overall} shows the attack performance of our training-based on seven datasets with different GNN methods. 
For most of the datasets, our attacks can achieve more than $65\%$ attack precision, which confirms our attacks are effective in various GNN models on different tasks.
Note that, for some of the datasets, such as DD and CIFAR10, our attacks achieve more than $0.85$ F1 Score, which means the target models are highly vulnerable to our proposed training-based MIAs. 
The results also depict that different GNN methods may have different levels of robustness against MIAs. 
It is observed that GateGCN and GCN are more vulnerable to inference attacks while GIN is more robust. 
It might be due to the different aggregation and transformation strategies used by these models. 
GIN aggregated via MLP can be more generalised to different input records, which increase the difficulty of membership inference.  

\noindent \textbf{Impact of attack strategies.} 
We evaluate how the size of the input vector affects our attacks. Section~\ref{subsec:dimension} mentions that the  dimensions of confidence vectors may be different for a cross-domain attack, so we distil them to vectors with lower dimensions. 
Table~\ref{tab:DifferentVectors} shows the comparison among the attacks on OGBG\_PPA, whose class is 37, with different sizes of confidence vectors used for attack models. 
It can be found that distilled vectors with 10 dimensions achieve higher attack rates compared to the others. 
Therefore, appropriately distilling the confidence vectors for attack model training can improve the attack performance. 

\noindent \textbf{Threshold-based attacks.} 
Table~\ref{tab:overall_threshold} demonstrates the attack performance of our threshold-based MIAs based on different matrix thresholds. 
Comparing training-based attacks, the threshold-based attacks are only effective via some of the thresholds while performing poorly via others. 
%

\noindent \textbf{Impact of the confidence metrics selection. }
We investigate how the confidence metrics affect our attacks. 
From the results in Table~\ref{tab:overall_threshold}, it is obvious that only the attacks using cross-entropy as the metric achieve satisfying performance. 
Other attacks perform much worse in most of the attack settings. 
The reason behind it is that only using the metrics for target model training can achieve high performance. 
As mentioned in Section~\ref{sec:method}, the intuition of our design is that the models will have higher confidence values for the members. 
However, based on our observation, this phenomenon only happens when the confidence value is measured by its loss function. 
So the attacks may perform badly when using other metrics. 

\begin{table*}[t]
\normalsize
\centering
\begin{tabular}{c|c|ccc}
  \hline
  Model  Architecture & Train-test Gap & Ordinary Vectors & Top-10 Vectors & Top-3 Vectors \\
  \hline
        28-layer & 0.412 & 0.826(0.011) & \textbf{0.839(0.009)} & 0.835(0.012) \\
        22-layer & 0.484 & 0.858(0.009) & 0.864(0.011) & \textbf{0.865(0.009)} \\
        20-layer & 0.480 & 0.860(0.013) & \textbf{0.873(0.012)} & 0.867(0.008) \\
        18-layer & 0.490 & \textbf{0.864(0.010)} & 0.862(0.012) & 0.859(0.010) \\
        16-layer & 0.510 & 0.852(0.012) & \textbf{0.863(0.012)} & 0.861(0.011) \\
        12-layer & 0.500 & 0.838(0.010) & 0.841(0.007) & \textbf{0.844(0.011)} \\
  \hline
\end{tabular}
\caption{F1 scores comparison among attack models with different input vectors. The best results are highlighted. }
\label{tab:DifferentVectors}
\vspace{-5pt}
\end{table*}


\subsubsection{Comparison to Non-graph-structure Model}
We now compare our attacks on GNNs with the MLP baseline as~\cite{abs-2003-00982}. 

\begin{framed}
    \noindent \textbf{Findings \#2:} 
    \textit{Comparing with MLP, graph-level GNN classification models are even more vulnerable to the MIAs in certain applications. }
\end{framed}
We consider the MLP baseline as the non-graph-structure ML method which simply updates the node independent as $h_i^{l+1} = \sigma(W^lh_i^l)$. 
From the results, our attacks reach better performance on GNNs than MLP for most of the datasets. 
Table~\ref{tab:overall} highlights the attacks with the best performance. It is shown that all the MLP models are more robust than GateGCN and GCN models; namely, some types of GNNs are more vulnerable to MIAs than MLPs. 
For example, our attacks targeting at GateGCN model trained on DD has a $0.885$ F1 score, which is about $0.333$ higher than $0.552$ for the MLP model. 

\subsubsection{Transferability}
\begin{figure*}[t]
    \centering
    \begin{minipage}[t]{0.3\textwidth}
        \centering
        \includegraphics[width=1\textwidth, height=0.75\textwidth]{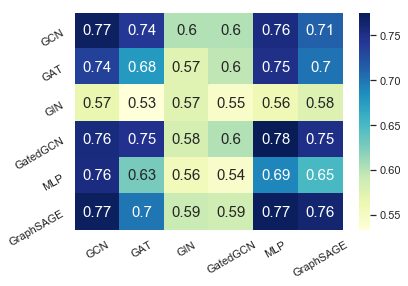}
     
    \end{minipage}
    \begin{minipage}[t]{0.3\textwidth}
        \centering
         \includegraphics[width=1\textwidth, height=0.75\textwidth]{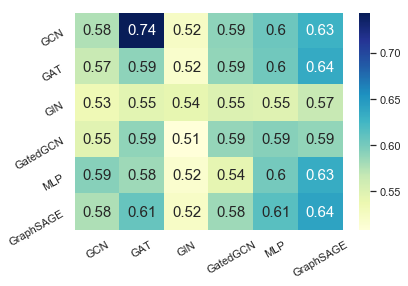}
    \end{minipage}
    \begin{minipage}[t]{0.3\textwidth}
         \centering
         \includegraphics[width=1\textwidth, height=0.75\textwidth]{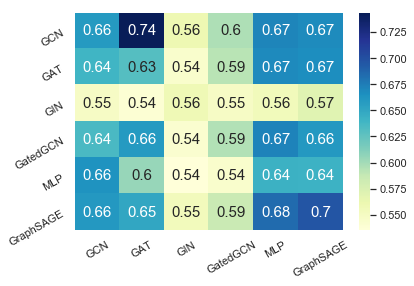}
    \end{minipage}
    \caption{Confusion matrix of the attack performance for using different GNN methods (left is the attack precision, the middle is the attack recall and right is the F1 score).}
    \label{fig:confu_matrix_GNN}
\end{figure*}

Finally, we explore the transferability of our attacks by relaxing the assumptions of the attacker's background knowledge about the target models and shadow dataset. 
\begin{framed}
    \noindent \textbf{Findings \#3:} 
    \textit{Our MIAs have strong transferability. The attacks are still effective with the shadow models trained as different GNN types and different datasets, or without training the shadow models. }
\end{framed}
\noindent \textbf{Training-based attacks.} 
As discussed in Section~\ref{sec:problem}, the attacker is assumed to have some knowledge of the GNN methods of the target models. 
However, in real-world scenarios, he may not know what types of GNNs are used for training the target models. 
As a result, the transferability of our attacks when mismatching the GNN methods needs to be investigated. 
Figure~\ref{fig:confu_matrix_GNN} reports the confusion matrix of the attack performance for the targeted/shadow model using different GNN methods. 
It can be found that, most of our transferring attacks are effective with a reduction of F1 score within $0.1$.

\begin{figure*}[t]
    \centering
    \begin{minipage}[t]{0.3\textwidth}
        \centering
        \includegraphics[width=1\textwidth, height=0.75\textwidth]{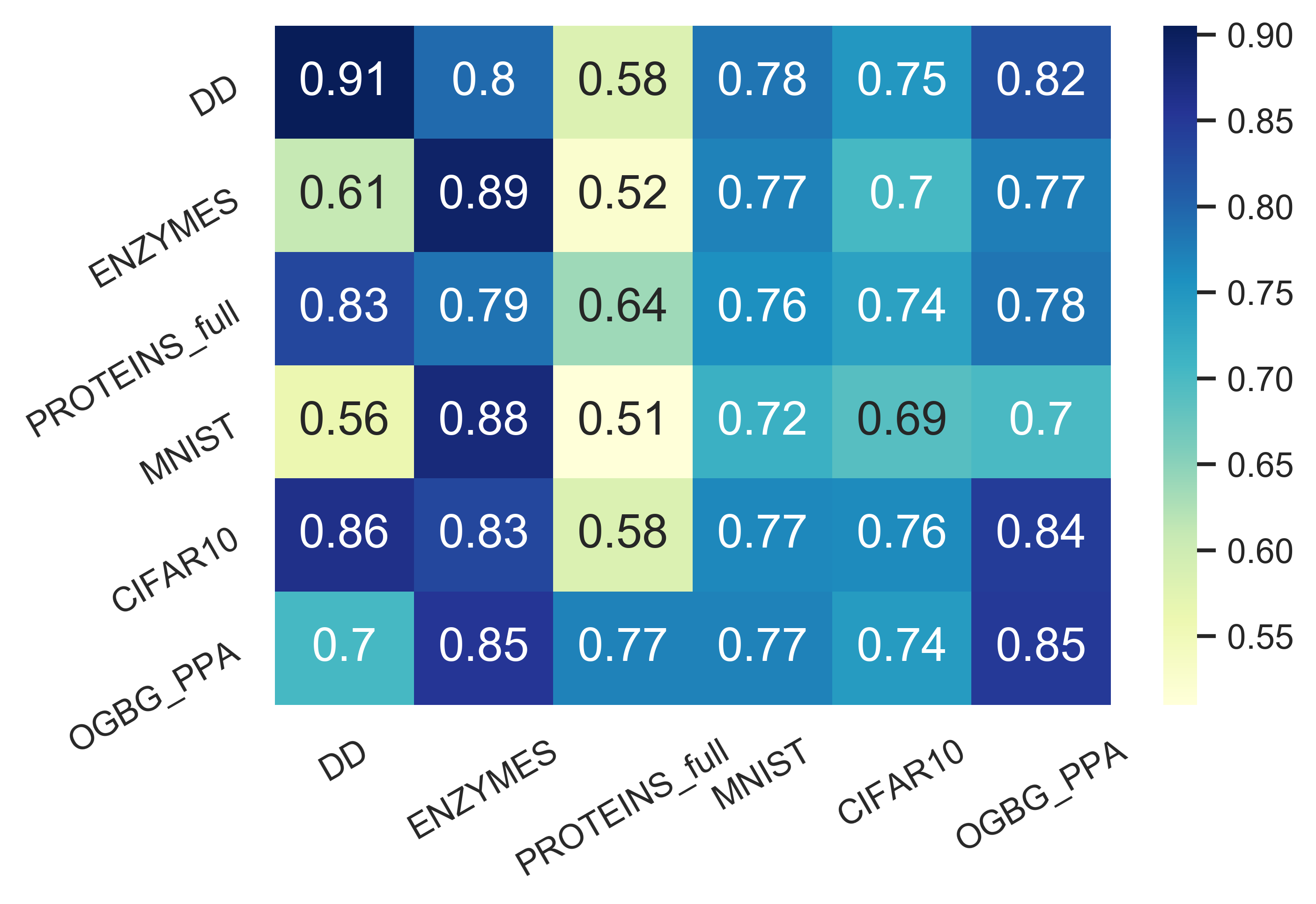}
    \end{minipage}
    \begin{minipage}[t]{0.3\textwidth}
        \centering
         \includegraphics[width=1\textwidth, height=0.75\textwidth]{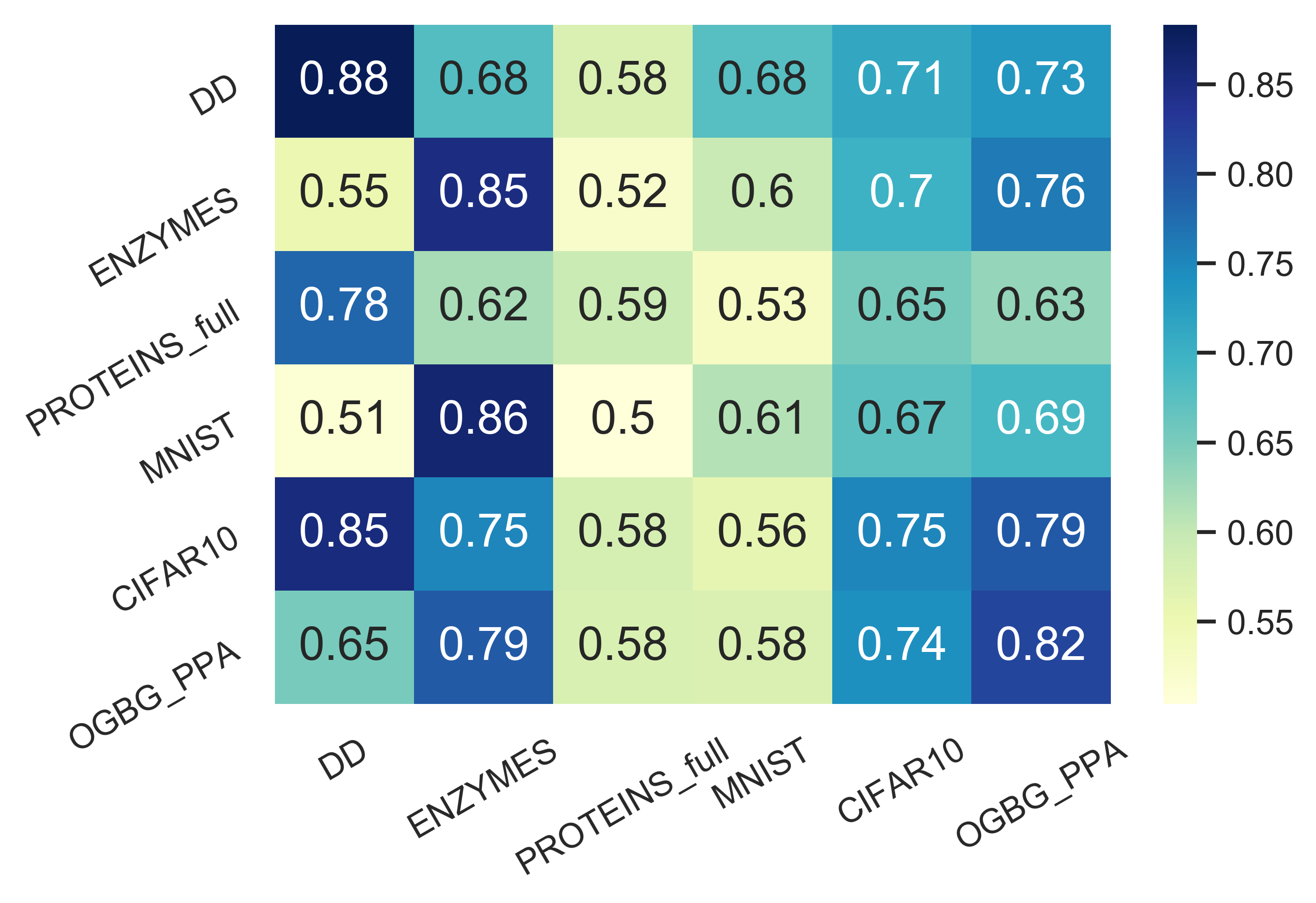}
    \end{minipage}
    \begin{minipage}[t]{0.3\textwidth}
         \centering
         \includegraphics[width=1\textwidth, height=0.75\textwidth]{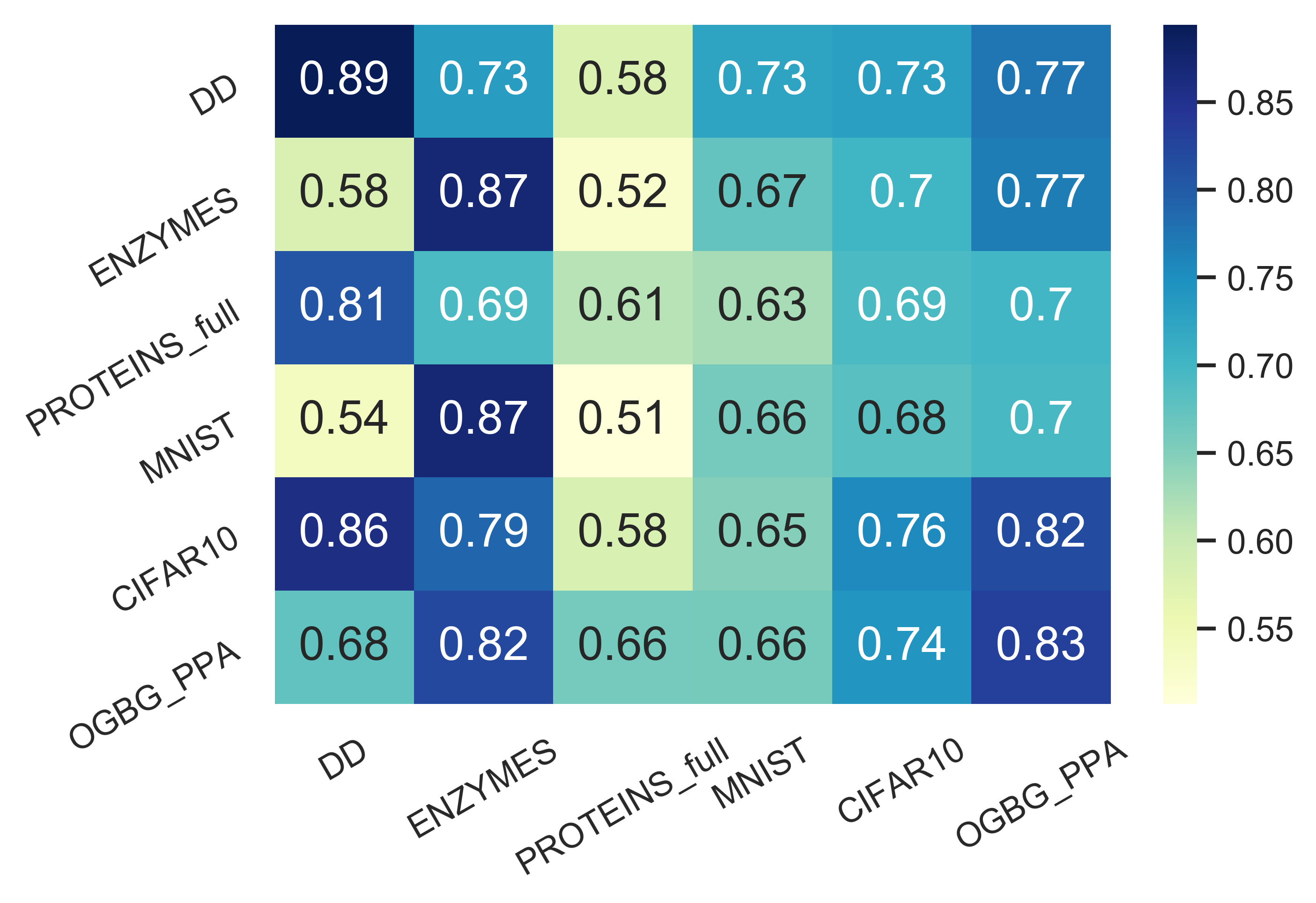}
    \end{minipage}
    \caption{Confusion matrix of the attack performance for using different shadow datasets (left is the attack precision, the middle is the attack recall and right is the F1 score).}
    \label{fig:confusion_matrix_dataset}
\end{figure*}

Besides the knowledge about the GNN methods, another important knowledge is the shadow dataset. %
In practice, it might be difficult for the attacker to gather shadow datasets in the same domain as the target. So, we evaluate our attacks' transferability in the cross-domain of shadow datasets. 
Figure~\ref{fig:confusion_matrix_dataset} shows the confusion matrix of the attack performance for the targeted/shadow model trained on different datasets. 
It can be found that our transferring attacks among cross-domain shadow datasets are still effective. 
As a result, the knowledge of the shadow dataset affects less in our attacks and our training-based MIAs are shown to have strong transferability. 

\noindent \textbf{Threshold-based attacks.} 
Different from the training-based attacks, the threshold-based attacks do not need to train shadow models.
As a result, the GNN methods of the target model do not affect the attacks and we only consider the transferability of our attacks for the cross-domain settings. 
The attackers should properly select the thresholds for their attacks based on the threshold distribution, which can be obtained by querying the target model with records in the same or similar domain. 
In practice, the optimal thresholds vary for data from different domains which places a higher demand on the knowledge of the attackers. 
We observe that the mismatching of the dataset domain can lead to a dramatic reduction of the attack effectiveness for the transferring attacks in cross-domain. 
As a result, the transferability of the threshold-based attacks is poor and it is hard for the attacker to apply the attacks without the same domain knowledge. 

\subsubsection{Comparison between Two Proposed Attacks}
After confirming both our training-based attacks and the threshold-based attacks can be effective when inferring the membership of the GNNs, we summarise the advantages and disadvantages for both of them.
\begin{framed}
    \noindent \textbf{Findings \#4:} 
    \textit{The threshold-based MIAs can achieve even better attack performance but much less transferability compared to the training-based attacks. }
\end{framed}
Firstly, the threshold-based attacks achieve the highest attack performance in most of the attack settings. 
In addition, there is no GNN method or dataset which is significantly more robust than others under the threshold-based attacks, while we observed that GIN is less vulnerable compared with training-based attacks against other types of GNNs. 
Furthermore, the threshold-based attacks require fewer computation resources since no attack models need to be trained during the attacks. 
However, the threshold-based attacks also have limitations. 
The selection of the thresholds is non-trivial, which may significantly affect the attack performance. 
Moreover, the transferability of the threshold-based attacks is much poorer than the training-based attacks, as the selection for the confidence metric is critical. 

\subsection{Factors Affecting Attack Performance}

\begin{table*}[t]
\normalsize
    \centering
    \begin{tabular}{c|c|c|ccc|cc}
    \hline
      \multicolumn{3}{c|}{}  & \multicolumn{3}{c|}{Target Graph Data Property} & \multicolumn{2}{c}{Target GNN Model Property} \\
    \cline{4-8}
      \multicolumn{3}{c|}{} & \#Nodes & \#Edges & Graph Density & \#Classes & Train-Test Gap \\
    \hline
        \multirow{4}{*}{Graph-level} & \multirow{2}{*}{Training-based} & GCN & -0.0296 & -0.0256 & -0.0155 & 0.8420 & 0.8110\\
                    &  & MLP & -0.1071 & -0.1125 & 0.1038 & 0.4748 & 0.5562 \\
    \cline{2-8}
                    &  \multirow{2}{*}{Threshold-based} & GCN & 0.0235 & 0.0268 & 0.0687  & 0.2385 & - \\
                    &  & MLP & 0.0001 & 0.0099 & 0.0846 & 0.3662 & - \\
    \hline
        \multirow{2}{*}{Node-level} & \cite{abs-2101-06570} & GCN  & - & - & - & 0.3857 & -0.2524 \\
                   & \cite{he2021node} & GraphSAGE & - & - & 0.7023 & -0.0550 & 0.2986 \\
    \hline
    \end{tabular}
    \caption{Correlations between several potential factors and attack performance.}
    \label{tab:statisticVSattacks}
\end{table*}

We have shown the effectiveness of our attacks under different attack settings. 
To fully understand the implications behind them, we further analyse our MIAs by adjusting several potential factors and discussing their effects on the attack performance. 
\subsubsection{Effect of the Target Model Property}
We first discuss how the target model property can affect the attack performance. 
%
%
\begin{framed}
    \noindent \textbf{Findings \#5:} 
    \textit{Overfitting is the most significant factor that affects the MIA performance on graph-level classification tasks. }
\end{framed}

\noindent \textbf{Impact of the overfitting. }
Similar to the prior works~\cite{ShokriSSS17,YeomGFJ18}, we first analyse the relationship between the overfitting level of the target GNN models and the attack performance. 
Overfitting represents how the model treats the input differently for members or non-member. 
Intuitively, the more difference the GNN models treat between the member and non-member input records, the easier our attack can be applied. 
Based on our experiments, it is also a significant factor that impacts our attack performance. 

\begin{figure}
    \centering
    \begin{subfigure}{\columnwidth}
        \includegraphics[width=\columnwidth, height=0.35\columnwidth]{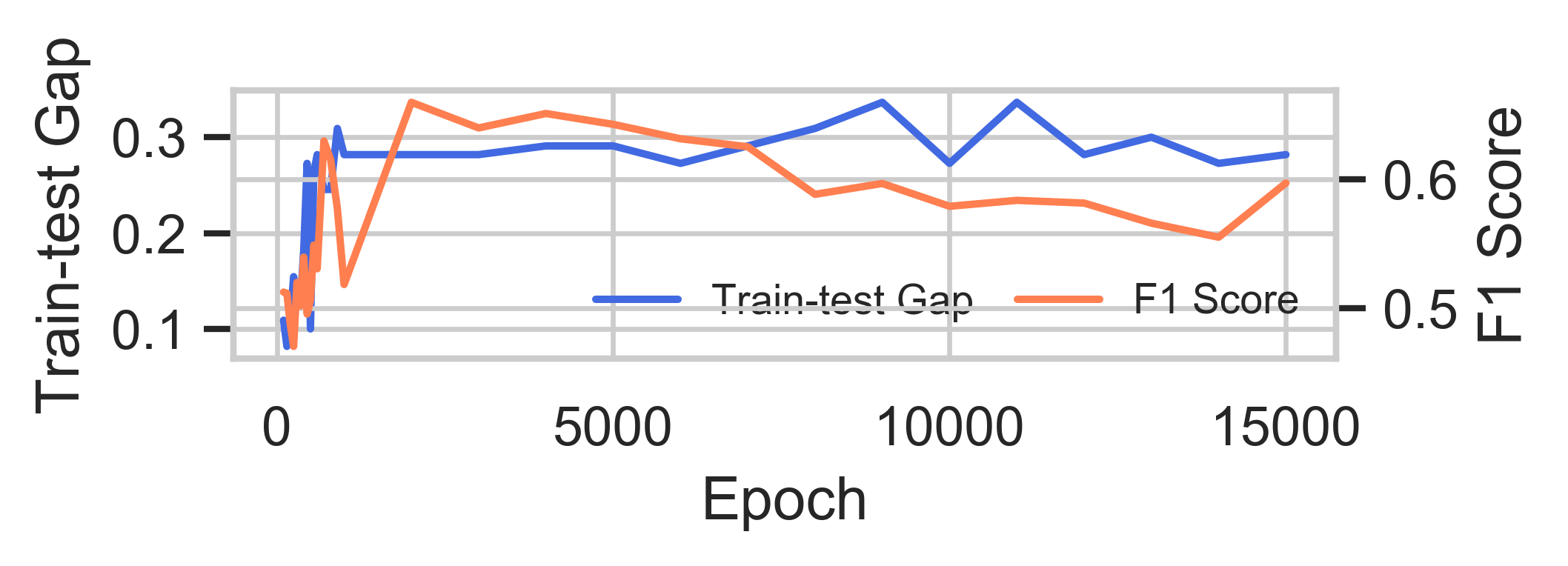} 
        \caption{PROTIEN\_full}
    \end{subfigure}
    \begin{subfigure}{\columnwidth}
        \includegraphics[width=\columnwidth, height=0.35\columnwidth]{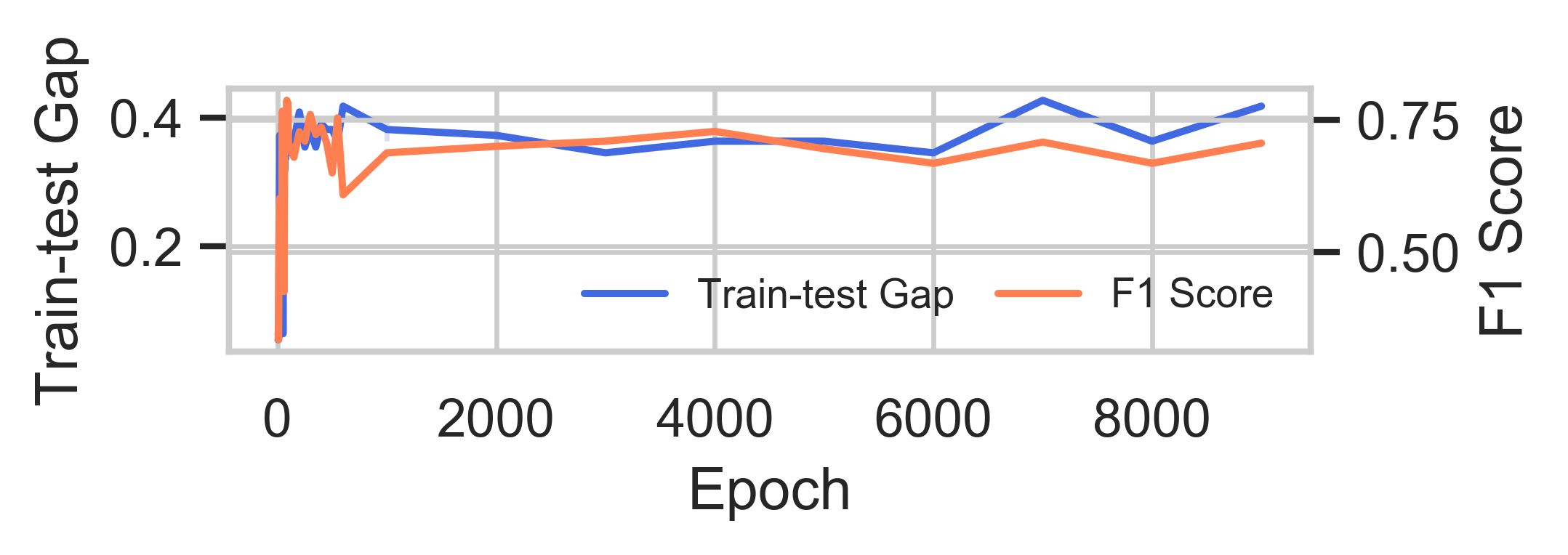}
        \caption{DD}
    \end{subfigure}
    \caption{Impact of the overfitting on training-based attacks.}
    \label{fig:overfitting}
\end{figure}

We measure the overfitting level by the train-test gap of the target model as previous MIA works~\cite{ShokriSSS17,YeomGFJ18}. 
Figure~\ref{fig:overfitting} shows how the train-test gap and the F1 score change when increasing the training epoch. 
Generally, after more epoch training, the target model becomes overfit and the train-test gap increases. 
Accordingly, the F1 scores of the attack also increase significantly which indicates that GNN models with higher overfitting levels are more vulnerable to the MIAs. 

\begin{figure}
    \centering
    \begin{subfigure}{\columnwidth}
        \includegraphics[width=\columnwidth, height=0.35\columnwidth]{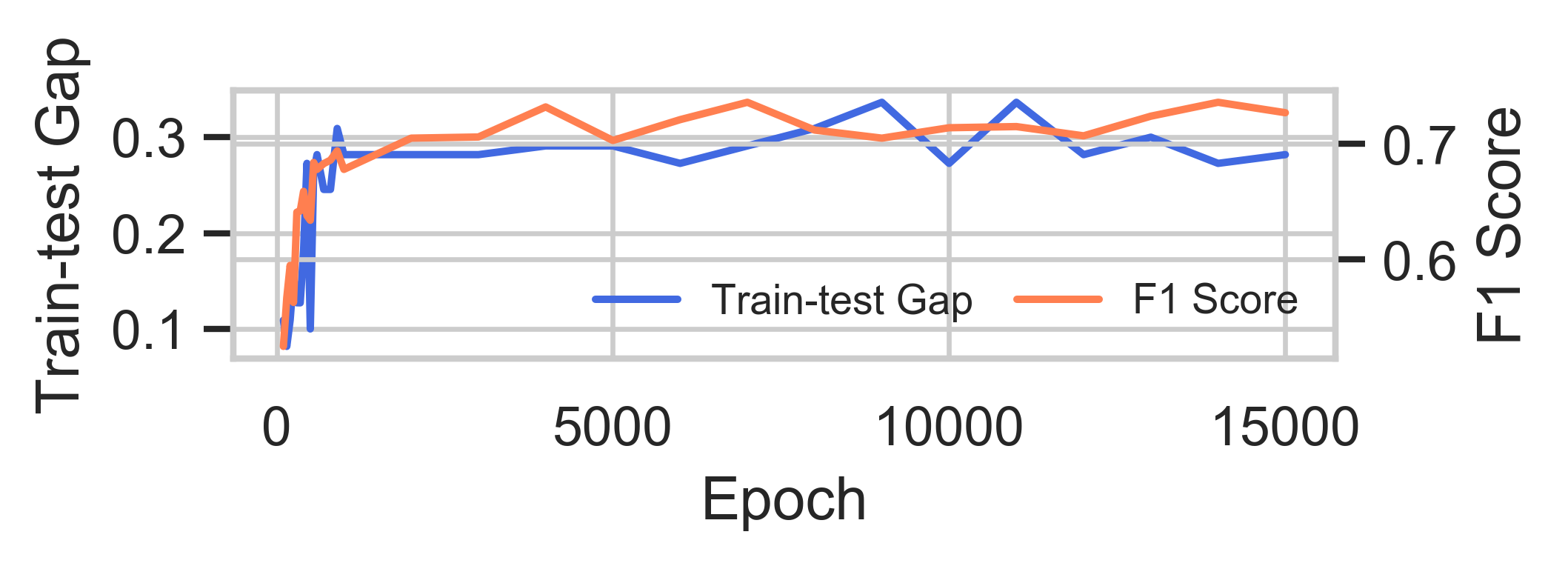} 
        \caption{PROTIEN\_full}
    \end{subfigure}
    \begin{subfigure}{\columnwidth}
        \includegraphics[width=\columnwidth, height=0.35\columnwidth]{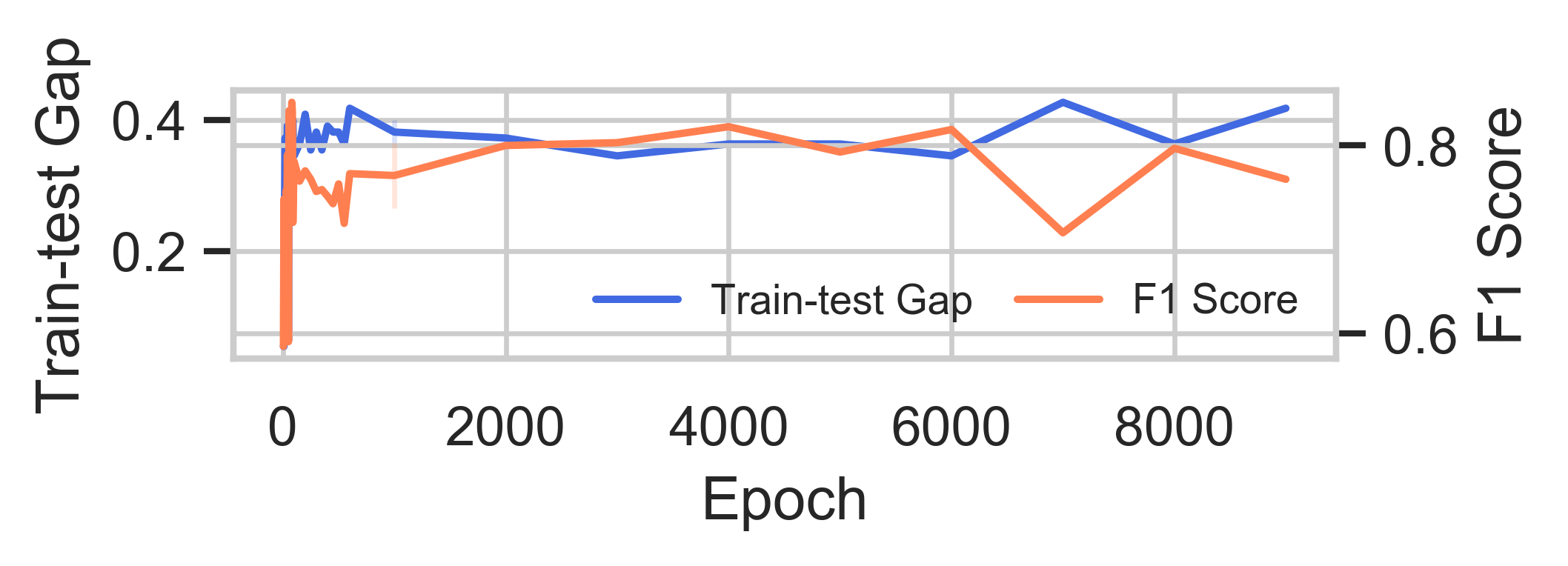}
        \caption{DD}
    \end{subfigure}
    \caption{Impact of the overfitting on threshold-based attacks. }
    \label{fig:overfitting_threshold}
\end{figure}

We evaluate how overfitting affects threshold-based attacks. 
Figure~\ref{fig:overfitting_threshold} illustrates the target model train-test gap and the attack performance for increasing training epochs. 
The results indicate that the threshold-based attacks are also positively correlated to the overfitting. 
In addition, comparing the training-based attacks, the threshold-based attacks are more sensitive to the fluctuation of the overfitting.  

\noindent \textbf{Impact of the model architecture complexity.}
We then evaluate how the model architecture affects our attacks. 
As known, the complexity of the model structure may affect the overfitting level of the model and then the attack performance. 
To eliminate the impacts from GNN aggregation methods, we evaluate it on DeepGCN~\cite{Li0TG19}, which is a large scale GNN with deeper layers. 
Table~\ref{tab:DifferentVectors} reports the relationship between the number of layers in the DeepGCN model and the effectiveness of the attacks. 
We observe that adding more layers to the DeepGCN model will reduce the train-test gaps and increase the F1 scores of our attacks. 
As mentioned in~\cite{Li0TG19}, adding layers may lead to higher training loss. 
In MIAs, higher training loss translates to less confidence in the member data which reduces the attack performance.  
Therefore, we can observe that the models with deeper layers achieve slightly stronger robustness. 


\subsubsection{Effect of the Target Graph Data Property}
%
We then explore how the property of the training data will affect the attack performance. 
Namely, whether some of the datasets are vulnerable or robust to the MIAs inherently.

\begin{framed}
    \noindent \textbf{Findings \#6:} 
    \textit{GNNs for graph classification with more classes or trained by graph data with larger average degrees are inherently more vulnerable to the MIAs. }
\end{framed}

We evaluate the correlation coefficient between the statistic values and the F1 score. 
The results of Spearman's correlations for both graph-level classification on PORITIEN\_full and node-level classification on CORA are shown in Table~\ref{tab:statisticVSattacks}. 
It can be found that for graph-level GNNs, the property related to the model such as the overfitting level (train-test gap) is highly related to the attack effectiveness, while the statistic of the training graph does not affect the attacks. 
However, node-level classification, \cite{abs-2101-06570} emphasises that the graph can significantly affect their attacks while the overfitting level of the target model does not. 
The experimental results in~\cite{he2021node} about the graph density also satisfy this observation as shown in Table~\ref{tab:statisticVSattacks}. 

The intuition behind this phenomenon is that the confidence vectors with a higher number of classes analysed by the attacker have higher dimensions and contain further information. 
Previous MIA works on images have also shown similar observations where MIAs are less effective when targeting binary classification tasks~\cite{ShokriSSS17,YeomGFJ18}. 
As a result, it is more challenging to design and train attack models with high accuracy that can identify the membership based on these confidence vectors with low dimensions. 
Besides, we found that the GNN models trained by the dataset with a higher average degree are more vulnerable to our MIAs, which might be caused by the aggregation strategies of GNNs. 
According to our observations, models trained on graphs with a larger average degree are more confident in the membership records. 
Consequently, such confidence in the membership records benefits the classification of the attack models and improves the attack performance. 
We consider exploring the rationale behind this observation as our future work.


%

\subsection{Comparison to MIAs on Node-level GNNs}
To investigate the difference between our MIAs on graph-level GNNs and prior works on node-level GNNs, we compare and discuss our above findings to theirs. 
Table~\ref{tab:statisticVSattacks} shows the comparison among the important effect factors of MIAs on different GNN applications. 
It can be found that graph-level MIAs are more correlated to the model property, while the performance of node-level MIAs depends more on the graph data property. 

This observation actually satisfied the implications behind the two attacks.  
Previous attack performance on inferring the membership of only one node has shown to be highly correlated with its neighbours ~\cite{abs-2101-06570,he2021node}. 
It is consistent with their attack design, where the attack model derives the membership based on the posterior of also the neighbouring nodes. 
Therefore, the intuition behind their attack is exploiting the correlation among the target nodes and their neighbours. 
On the contrary, our attacks infer the membership of the inputs based on the final posterior of the entire graph. 
The effectiveness of the attacks fully relies on how the target GNN model overfits its training graph data. 
Therefore, the correlation between the target model property, especially the overfitting level, becomes the most significant factor in MIAs on graph-level classification. 



\section{Conclusion and Future Directions}
\label{sec:conclusion}
This paper investigates how GNNs are vulnerable to the MIAs and develops training-based and threshold-based attacks against various target GNN models. 
The experiment results demonstrate that our attacks are effective against GNN models. We investigate several impact factors of the MIAs, which are common in other ML models or unique in GNNs. Our findings show that overfitting is still the most significant factor that affects the attack performance. To further evaluate the attack transferability, we apply and compare the attacks with different adversarial background knowledge on the shadow datasets and the GNN types. The results demonstrate that the attack can still be effective even the type of the target model is unknown. 

As a future direction, defence against the MIAs on GNNs becomes emerging. Moreover, how to deploy the MIAs utilising label-only exposures as~\cite{abs-2007-14321} can also be an interesting extension of our work. We leave them as future work.

\bibliographystyle{IEEEtran}
\bibliography{IEEEexample}

\begin{thebibliography}{10}
\providecommand{\url}[1]{#1}
\csname url@samestyle\endcsname
\providecommand{\newblock}{\relax}
\providecommand{\bibinfo}[2]{#2}
\providecommand{\BIBentrySTDinterwordspacing}{\spaceskip=0pt\relax}
\providecommand{\BIBentryALTinterwordstretchfactor}{4}
\providecommand{\BIBentryALTinterwordspacing}{\spaceskip=\fontdimen2\font plus
\BIBentryALTinterwordstretchfactor\fontdimen3\font minus
  \fontdimen4\font\relax}
\providecommand{\BIBforeignlanguage}[2]{{%
\expandafter\ifx\csname l@#1\endcsname\relax
\typeout{** WARNING: IEEEtran.bst: No hyphenation pattern has been}%
\typeout{** loaded for the language `#1'. Using the pattern for}%
\typeout{** the default language instead.}%
\else
\language=\csname l@#1\endcsname
\fi
#2}}
\providecommand{\BIBdecl}{\relax}
\BIBdecl

\bibitem{XuHXXXY20}
H.~Xu, C.~Huang, Y.~Xu, L.~Xia, H.~Xing, and D.~Yin, ``Global context enhanced
  social recommendation with hierarchical graph neural networks,'' in
  \emph{Proc. {IEEE} {ICDM}}, 2020.

\bibitem{WangYDHLN19}
W.~Wang, H.~Yin, X.~Du, W.~Hua, Y.~Li, and Q.~V.~H. Nguyen, ``Online user
  representation learning across heterogeneous social networks,'' in
  \emph{Proc. {ACM SIGIR}}, 2019.

\bibitem{wu2020comprehensive}
Z.~Wu, S.~Pan, F.~Chen, G.~Long, C.~Zhang, and S.~Y. Philip, ``A comprehensive
  survey on graph neural networks,'' \emph{IEEE transactions on neural networks
  and learning systems}, vol.~32, no.~1, pp. 4--24, 2020.

\bibitem{liu2021graph}
Y.~Liu, S.~Pan, M.~Jin, C.~Zhou, F.~Xia, and P.~S. Yu, ``Graph self-supervised
  learning: A survey,'' \emph{arXiv:2103.00111}, 2021.

\bibitem{ChenYZF20}
D.~Chen, N.~Yu, Y.~Zhang, and M.~Fritz, ``Gan-leaks: {A} taxonomy of membership
  inference attacks against generative models,'' in \emph{Proc. {ACM CCS}},
  2020.

\bibitem{ZhuLLBZZL20}
Y.~Zhu, X.~Luo, Y.~Li, B.~Bu, K.~Zhou, W.~Zhang, and M.~Lu, ``Heterogeneous
  mini-graph neural network and its application to fraud invitation
  detection,'' in \emph{Proc. {IEEE} {ICDM}}, 2020.

\bibitem{PangZJLW20}
R.~Pang, X.~Zhang, S.~Ji, X.~Luo, and T.~Wang, ``Advmind: Inferring adversary
  intent of black-box attacks,'' in \emph{Proc. {ACM KDD}}, 2020.

\bibitem{263820}
X.~He, J.~Jia, M.~Backes, N.~Z. Gong, and Y.~Zhang, ``Stealing links from graph
  neural networks,'' in \emph{Proc. {USENIX} Security}, 2021.

\bibitem{GongL18}
N.~Z. Gong and B.~Liu, ``Attribute inference attacks in online social
  networks,'' \emph{{ACM} Trans. Priv. Secur.}, vol.~21, no.~1, pp. 3:1--3:30,
  2018.

\bibitem{abs-2007-14321}
C.~A. Choquette{-}Choo, F.~Tram{\`{e}}r, N.~Carlini, and N.~Papernot,
  ``Label-only membership inference attacks,'' \emph{CoRR}, vol.
  abs/2007.14321, 2020.

\bibitem{demystifying19}
S.~Truex, L.~Liu, M.~Gursoy, L.~Yu, and W.~Wei, ``Demystifying membership
  inference attacks in machine learning as a service,'' \emph{IEEE transactions
  on services computing}, pp. 1--1, 2019.

\bibitem{Salem0HBF019}
A.~Salem, Y.~Zhang, M.~Humbert, P.~Berrang, M.~Fritz, and M.~Backes,
  ``Ml-leaks: Model and data independent membership inference attacks and
  defenses on machine learning models,'' in \emph{Proc. {NDSS}}, 2019.

\bibitem{ShokriSSS17}
R.~Shokri, M.~Stronati, C.~Song, and V.~Shmatikov, ``Membership inference
  attacks against machine learning models,'' in \emph{Proc. {IEEE} {S\&P}},
  2017.

\bibitem{abs-2101-06570}
I.~E. Olatunji, W.~Nejdl, and M.~Khosla, ``Membership inference attack on graph
  neural networks,'' \emph{CoRR}, vol. abs/2101.06570, 2021.

\bibitem{he2021node}
X.~He, R.~Wen, Y.~Wu, M.~Backes, Y.~Shen, and Y.~Zhang, ``Node-level membership
  inference attacks against graph neural networks,'' \emph{arXiv preprint
  arXiv:2102.05429}, 2021.

\bibitem{YeomGFJ18}
S.~Yeom, I.~Giacomelli, M.~Fredrikson, and S.~Jha, ``Privacy risk in machine
  learning: Analyzing the connection to overfitting,'' in \emph{Proc. {IEEE}
  {CSF}}, 2018.

\bibitem{LeinoF20}
K.~Leino and M.~Fredrikson, ``Stolen memories: Leveraging model memorization
  for calibrated white-box membership inference,'' in \emph{Proc. {USENIX}
  Security}, 2020.

\bibitem{InferAttack22}
Z.~Zhang, M.~Chen, M.~Backes, Y.~Shen, and Y.~Zhang, ``Inference attacks
  against graph neural networks,'' in \emph{Proc. {USENIX} Security}, 2022.

\bibitem{HuangXYKAGM13}
J.~Huang, Y.~Xie, F.~Yu, Q.~Ke, M.~Abadi, E.~Gillum, and Z.~M. Mao,
  ``Socialwatch: detection of online service abuse via large-scale social
  graphs,'' in \emph{Proc. {ACM} {AsiaCCS}}, 2013.

\bibitem{GongFM14}
N.~Z. Gong, M.~Frank, and P.~Mittal, ``Sybilbelief: {A} semi-supervised
  learning approach for structure-based sybil detection,'' \emph{{IEEE} Trans.
  Inf. Forensics Secur.}, vol.~9, no.~6, pp. 976--987, 2014.

\bibitem{abs-1908-02591}
M.~Weber, G.~Domeniconi, J.~Chen, D.~K.~I. Weidele, C.~Bellei, T.~Robinson, and
  C.~E. Leiserson, ``Anti-money laundering in bitcoin: Experimenting with graph
  convolutional networks for financial forensics,'' \emph{CoRR}, vol.
  abs/1908.02591, 2019.

\bibitem{WangJG19}
B.~Wang, J.~Jia, and N.~Z. Gong, ``Graph-based security and privacy analytics
  via collective classification with joint weight learning and propagation,''
  in \emph{Proc. {NDSS}}.\hskip 1em plus 0.5em minus 0.4em\relax The Internet
  Society, 2019.

\bibitem{KongY13}
D.~Kong and G.~Yan, ``Discriminant malware distance learning on structural
  information for automated malware classification,'' in \emph{Proc. {ACM
  KDD}}, 2013.

\bibitem{NikolopoulosP17}
S.~D. Nikolopoulos and I.~Polenakis, ``A graph-based model for malware
  detection and classification using system-call groups,'' \emph{J. Comput.
  Virol. Hacking Tech.}, vol.~13, no.~1, pp. 29--46, 2017.

\bibitem{HassenC17}
M.~Hassen and P.~K. Chan, ``Scalable function call graph-based malware
  classification,'' in \emph{Proc. {ACM} {CODASPY}}, 2017.

\bibitem{YanYJ19}
J.~Yan, G.~Yan, and D.~Jin, ``Classifying malware represented as control flow
  graphs using deep graph convolutional neural network,'' in \emph{Proc. {IEEE}
  {DSN}}, 2019.

\bibitem{abs-1709-03741}
J.~Li, D.~Cai, and X.~He, ``Learning graph-level representation for drug
  discovery,'' \emph{CoRR}, vol. abs/1709.03741, 2017.

\bibitem{JiaSBZG19}
J.~Jia, A.~Salem, M.~Backes, Y.~Zhang, and N.~Z. Gong, ``Memguard: Defending
  against black-box membership inference attacks via adversarial examples,'' in
  \emph{Proc. {ACM CCS}}, 2019.

\bibitem{wang2019dgl}
M.~Wang, D.~Zheng, Z.~Ye, Q.~Gan, M.~Li, X.~Song, J.~Zhou, C.~Ma, L.~Yu,
  Y.~Gai, T.~Xiao, T.~He, G.~Karypis, J.~Li, and Z.~Zhang, ``Deep graph
  library: A graph-centric, highly-performant package for graph neural
  networks,'' \emph{arXiv preprint arXiv:1909.01315}, 2019.

\bibitem{abs-2003-00982}
V.~P. Dwivedi, C.~K. Joshi, T.~Laurent, Y.~Bengio, and X.~Bresson,
  ``Benchmarking graph neural networks,'' \emph{CoRR}, vol. abs/2003.00982,
  2020.

\bibitem{dobson2003distinguishing}
P.~D. Dobson and A.~J. Doig, ``Distinguishing enzyme structures from
  non-enzymes without alignments,'' \emph{Journal of molecular biology}, 2003.

\bibitem{SzklarczykGLJWH19}
D.~Szklarczyk, A.~L. Gable, D.~Lyon, A.~Junge, S.~Wyder, J.~Huerta{-}Cepas,
  M.~Simonovic, N.~T. Doncheva, J.~H. Morris, P.~Bork, L.~J. Jensen, and C.~von
  Mering, ``{STRING} v11: protein-protein association networks with increased
  coverage, supporting functional discovery in genome-wide experimental
  datasets,'' \emph{Nucleic Acids Res.}, vol.~47, no. Database-Issue, pp.
  D607--D613, 2019.

\bibitem{tiny_images}
A.~Krizhevsky, ``Learning multiple layers of features from tiny images,''
  \emph{University of Toronto}, 05 2012.

\bibitem{726791}
Y.~{Lecun}, L.~{Bottou}, Y.~{Bengio}, and P.~{Haffner}, ``Gradient-based
  learning applied to document recognition,'' \emph{Proceedings of the IEEE},
  vol.~86, no.~11, pp. 2278--2324, 1998.

\bibitem{AchantaSSLFS12}
R.~Achanta, A.~Shaji, K.~Smith, A.~Lucchi, P.~Fua, and S.~S{\"{u}}sstrunk,
  ``{SLIC} superpixels compared to state-of-the-art superpixel methods,''
  \emph{{IEEE} Trans. Pattern Anal. Mach. Intell.}, vol.~34, no.~11, pp.
  2274--2282, 2012.

\bibitem{abs-2007-08663}
C.~Morris, N.~M. Kriege, F.~Bause, K.~Kersting, P.~Mutzel, and M.~Neumann,
  ``Tudataset: {A} collection of benchmark datasets for learning with graphs,''
  \emph{CoRR}, vol. abs/2007.08663, 2020.

\bibitem{KipfW17}
T.~N. Kipf and M.~Welling, ``Semi-supervised classification with graph
  convolutional networks,'' in \emph{Proc. {ICLR}}.\hskip 1em plus 0.5em minus
  0.4em\relax OpenReview.net, 2017.

\bibitem{abs-1711-07553}
X.~Bresson and T.~Laurent, ``Residual gated graph convnets,'' \emph{CoRR}, vol.
  abs/1711.07553, 2017.

\bibitem{XuHLJ19}
K.~Xu, W.~Hu, J.~Leskovec, and S.~Jegelka, ``How powerful are graph neural
  networks?'' in \emph{Proc. {ICLR}}, 2019.

\bibitem{VelickovicCCRLB18}
P.~Velickovic, G.~Cucurull, A.~Casanova, A.~Romero, P.~Li{\`{o}}, and
  Y.~Bengio, ``Graph attention networks,'' in \emph{Proc. {ICLR}}, 2018.

\bibitem{HamiltonYL17}
W.~L. Hamilton, Z.~Ying, and J.~Leskovec, ``Inductive representation learning
  on large graphs,'' in \emph{Proc. {NIPS}}, 2017.

\bibitem{Li0TG19}
G.~Li, M.~M{\"{u}}ller, A.~K. Thabet, and B.~Ghanem, ``Deepgcns: Can gcns go as
  deep as cnns?'' in \emph{Proc. {ICCV}}, 2019.

\end{thebibliography}

\end{document}